%% file: 2022_emnlp_interactive_tm_label.tex
\pdfoutput=1

\documentclass[11pt]{article}

\usepackage{style/emnlp2023}

\usepackage{times}
\usepackage{latexsym}

\usepackage[T1]{fontenc}

\usepackage[utf8]{inputenc}

\usepackage{microtype}

\usepackage{inconsolata}

\title{Labeled Interactive Topic Models}

\author{Kyle Seelman \\
  Computer Science  \\
  University of Maryland \\
  \texttt{kseelman@umd.edu} \\ \And
  Mozhi Zhang \\
  Computer Science  \\
  Unviersity of Maryland  \\
  \texttt{mozhi@umd.edu} \\ \And
  Jordan Boyd-Graber \\
  \abr{cs}, iSchool, \abr{lcs}, \abr{umiacs} \\
  University of Maryland \\
 \texttt{jbg@umiacs.umd.edu}\\}

\newif\ifcomment\commenttrue
\input{style/preamble}

\begin{document}
\maketitle
\begin{abstract}
\input{sections/00-abstract}

\end{abstract}

\section{Topic Models Need Help} \label{sec:sections/10-intro} \input{sections/10-intro}
\section{The Best of Both Worlds: Neural Word Knowledge and Bayesian Informative Priors} \label{sec:sections/20-background} \input{sections/20-background} \label{background}
\section{Interactive Neural Topic Modeling} \label{sec:sections/30-methods} \input{sections/30-methods} \label{section:method}
\section{I-NTM Experimental Results } \label{sec:sections/40-experiments} \input{sections/40-experiments} \label{section:findings}
\section{Related Work} \label{sec:sections/60-related} \input{sections/60-related}
\section{Conclusion and Future Work} \label{sec:sections/70-conclusion} \input{sections/70-conclusion}

\clearpage
\section*{Limitations} \label{sec:sections/80-limitations} \input{sections/80-limitations} \label{limits}
\section*{Ethical Considerations} \label{sec:sections/90-ethics} \input{sections/90-ethics}

\bibliography{bib/journal-full,bib/anthology, bib/custom, bib/jbg}
\bibliographystyle{acl_natbib}

\appendix
\section{Datasets} \label{sec:sections/100-appendix} \input{sections/100-appendix}

\end{document}

%% file: style/preamble.tex
\usepackage[a-1b]{pdfx}

\usepackage{framed}
\usepackage{mdwlist}
\usepackage{siunitx}
\usepackage{latexsym}
\usepackage{colortbl}
\usepackage{xcolor}
\usepackage{nicefrac}
\usepackage{booktabs}
\usepackage{fnpct}
\usepackage{amsfonts}
\usepackage[T1]{fontenc}
\usepackage{bold-extra}
\usepackage{amsmath}
\usepackage{amssymb}
\usepackage{bm}
\usepackage{graphicx}
\usepackage{mathtools}
\usepackage{microtype}
\usepackage{multirow}
\usepackage{multicol}
\usepackage{xpatch}
\usepackage{latexsym,comment}
\usepackage[normalem]{ulem}

\newcommand*{\missingreference}{{\Huge \colorbox{red}{?reference?}}}
\newcommand*{\missingcitation}{{\Huge \colorbox{red}{?citation?}}}

\makeatletter
\xpatchcmd{\@setref}{\bfseries}{\missingreference}{}{}
\def\@citex[#1]#2{\leavevmode
    \let\@citea\@empty
    \@cite{\@for\@citeb:=#2\do
        {\@citea\def\@citea{,\penalty\@m\ }%
            \edef\@citeb{\expandafter\@firstofone\@citeb\@empty}%
            \if@filesw\immediate\write\@auxout{\string\citation{\@citeb}}\fi
            \@ifundefined{b@\@citeb}{\hbox{\reset@font\missingcitation}%
                \G@refundefinedtrue
                \@latex@warning
                {Citation `\@citeb' on page \thepage \space undefined}}%
            {\@cite@ofmt{\csname b@\@citeb\endcsname}}}}{#1}}
\makeatother

\newcommand{\gem}[1]{\mbox{\textsc{gem}}}
\newcommand{\abr}[1]{\textsc{#1}}

\newcommand{\g}{\, | \,}

\newcommand{\hidetext}[1]{}
\newcommand{\ignore}[1]{}

\ifcomment
    \newcommand{\pinaforecomment}[3]{\colorbox{#1}{\parbox{.8\linewidth}{#2: #3}}}

    \newcommand{\prtodo}[1]{\pinaforecomment{lightblue}{pr}{#1}}
    \newcommand{\prtodoi}[1]{\pinaforecomment{lightblue}{pr}{#1}}
\else
    \newcommand{\pinaforecomment}[3]{}
    \newcommand{\prtodo}[1]{}
    \newcommand{\prtodoi}[1]{}
\fi

\newcommand{\smallurl}[1]{ \begin{tiny}\url{#1}\end{tiny}}

\definecolor{lightblue}{HTML}{3cc7ea}
\definecolor{CUgold}{HTML}{CFB87C}
\definecolor{grey}{rgb}{0.95,0.95,0.95}
\definecolor{ceil}{rgb}{0.57, 0.63, 0.81}
\definecolor{UMDred}{HTML}{ed1c24}
\definecolor{UMDyellow}{HTML}{ffc20e}


%% file: sections/00-abstract.tex
Topic models are valuable for understanding extensive document collections, but they don't always identify the most relevant topics. Classical probabilistic and anchor-based topic models offer interactive versions that allow users to guide the models towards more pertinent topics. However, such interactive features have been lacking in neural topic models.
To correct this lacuna, we introduce a user-friendly interaction for neural topic models. This interaction permits users to assign a word label to a topic, leading to an update in the topic model where the words in the topic become closely aligned with the given label.
Our approach encompasses two distinct kinds of neural topic models. The first includes models where topic embeddings are trainable and evolve during the training process. The second kind involves models where topic embeddings are integrated post-training, offering a different approach to topic refinement.
To facilitate user interaction with these neural topic models, we have developed an interactive interface. This interface enables users to engage with and re-label topics as desired. We evaluate our method through a human study, where users can relabel topics to find relevant documents. Using our method, user labeling improves document rank scores, helping to find more relevant documents to a given query when compared to no user labeling.

%% file: sections/10-intro.tex
\begin{table*}[th]
    \centering
\begin{tabular}{ |p{6cm}||p{6cm}|  }
 \hline
 \multicolumn{2}{|c|}{\textbf{Topic}: Dengue outbreak in Asia} \\
 \multicolumn{2}{|c|}{\textbf{Query}: What countries are seeing an outbreak?} \\
 \hline
 No topic labeling & After topic labeling\\
 \hline
 \underline{Topic~0}: `dengue', `vaccine', `sanofi', `dengvaxia, `phillipines', `vaccination' & \textcolor{gray}{\underline{Topic~0}: `dengue', `vaccine', `sanofi', `dengvaxia', `phillipines', `vaccination'} \\
 \underline{Topic~1}: `virus', `countries', `new', `according', `dr', `pandemic' &  \textcolor{gray}{\underline{Topic~1}: `virus', `countries', `new', `according', `dr', `pandemic'} \\ 
 \underline{Topic~2}: `time', `get', `however', `gonaives', `haiti', `town, `stud' & \textbf{\underline{Topic~2}: `india', `genotype', `denv', `asian', `study', `singapore'}\\
 \hline
\end{tabular}
\caption{This figure demonstrates the capability of interactive topic modeling in refinign topics. Initially, `Topic 2' does not align with the user's requirements. (Left) Before the intervention, the closest words to the topic embeddings, as generated by the ETM, show that while the first two topics correlate with the task, `Topic 2' is unrelated. (Right)  The updated 'Topic 2' now closely aligns with the user-specified label, `india', showcasing how \abr{i-ntm} adapts in real-time to user input, giving greater relevance and accuracy in topic representation.}
\label{tab:example}
\end{table*}

Topic modeling is an unsupervised machine learning method for
analyzing a set of documents to learn meaningful clusters of related
words~\citep{boyd-graber-17}.
Despite decades of new models that purport to improve upon it, the
most popular method remains Latent Dirichlet
Allocation~\citep[\abr{lda}]{Blei2003LatentDA}, which is two decades
old.

This venerable model is still the workhorse for those who use
unsupervised analysis to discover the structure of document
collections in digital humanities~\citep{meeks2012digital},
bioinformatics~\citep{liu2016overview}, political
science~\citep{grimmer2013text}, and social
science~\citep{ramage_social_tm}.  
However, if you look at the computer science literature, topic
modeling has been taken over by neural
approaches~\citep{zhao2021topic}, such as the embedded topic model
(\abr{etm}) and contextualized topic models (\abr{ctm}). We review
\abr{lda} and neural topic models in Section~\ref{sec:review}.

So what explains this discrepancy?  
A sceptic would posit that there is not sufficient evidence to support
the claims that neural topic models are substantially better either in
terms of runtime, ease-of-use, or on human-centric
methods~\citep{hoyle-21}.
In addition to these legitimate concerns, there are also functional
lacunae: abilities ``classic'' topic models have that neural models
lack.
Neural models are often a ``take it or leave it'' proposition: if the
results do not match what you want, a user (particularly a non-expert
in machine learning) has little recourse.

In contrast, the probabilistic topic modeling literature has a rich
menu of options to improve topic models: incorporating rich priors~\citep{blei_super},
incorporating syntactic information~\citep{blei_syntax}, or structural priors from
covariates~\citep{jbg_syntax}.

Unfortunately, these improvements are not currently available for
neural topic models.

Making neural models interactive requires two things: models to support
interactivity and an interface to allow users to make changes to the model.
This paper provides both and applies them both to models by directly updating
topic embeddings (\abr{etm}) and by backpropagating downstream labels into
representations used by neural variational document model
(\abr{nvdm})\citealp{miao2016neural} and \abr{ctm}.

To use \abr{i-ntm} interactively---based on the topic label from the
user---we embed the label in the embedding space and \emph{move} the
corresponding topic embedding closer to the label. We detail the two different types of "moving" in Section~\ref{section:adjust}.
This adjusts the center of the topic embedding: throwing out unrelated
words, prioritizing words that are ``close'' to the users' label.
While there have been many previous works for interactive labeling,
our work introduces a way of improving topics through a natural way of
labeling that is typically done a posteriori and we provide a easy to
use interface for this.
We call this method \textit{Interactive Neural Topic Modeling} or
\abr{i-ntm}. Additionally, we provide a user-friendly interface to see the
documents clustered to each topic and allow the users to label topics as
desired.
As seen in the original paper and confirmed through automatic metrics on our
datasets, \abr{ctm} is overall the best neural topic model in terms of topic coherence and diversity. 

To demonstrate the efficacy of our interactive labeling method and interface,
we conduct a human study using the \abr{ctm} backend of \abr{i-ntm}. \abr{ctm}
was chosen since it showed to find the most diverse and coherent topics out of the 
three models we provide support for.
Additionally, if a user has a specific task when running a topic model on a
corpus, our interactive labeling method qualitatively helps users quickly
identify documents relevant to their information needs, as we will see in Section~\ref{section:findings}

%% file: sections/20-background.tex
\label{sec:review}

\begin{figure*}[t]
    \centering
    \includegraphics[width=\textwidth]{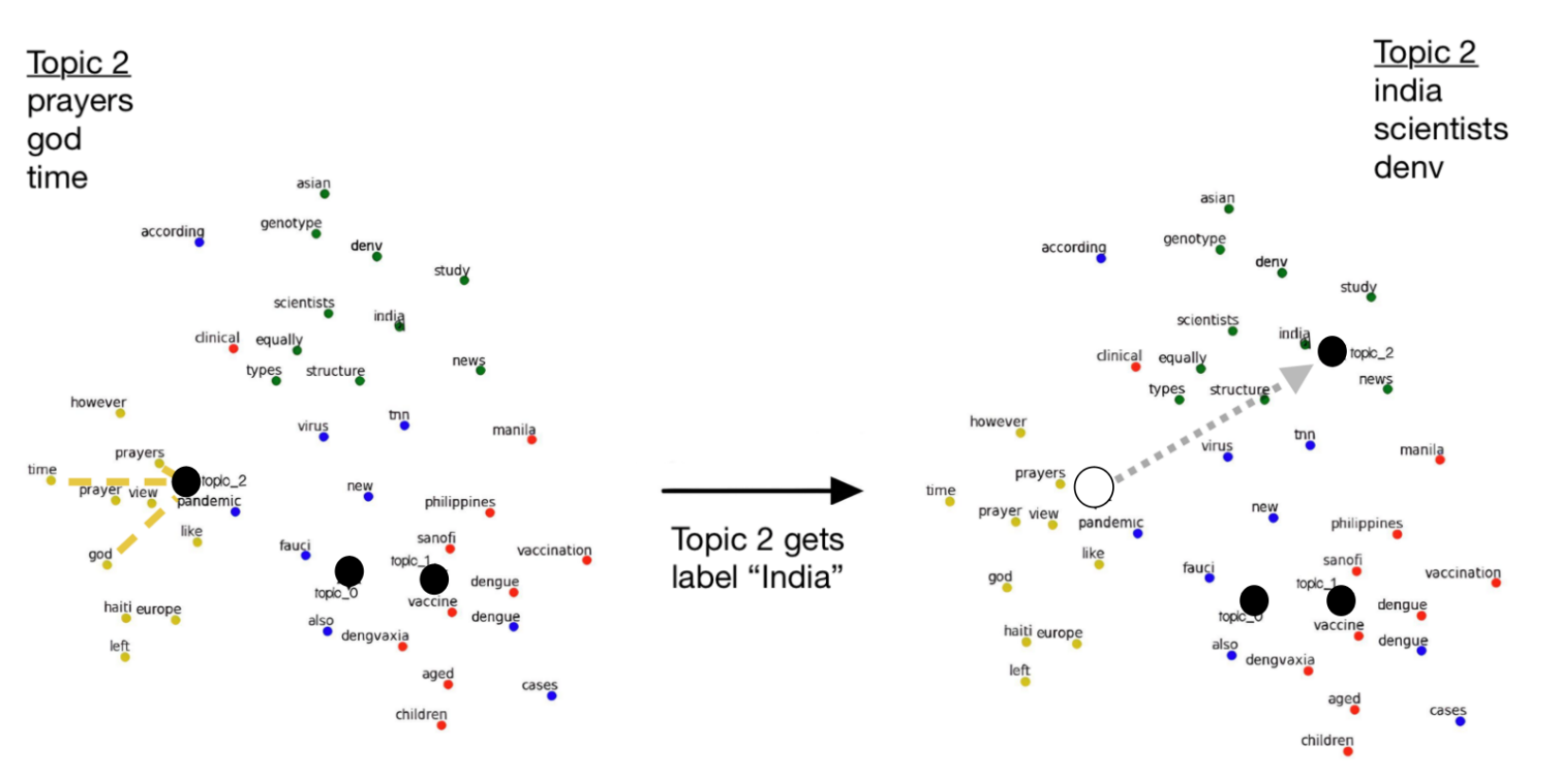}\hfill
    \caption{Visual representation labeling a new topic with out method, like in Table~\ref{tab:example}. Our method moves the embedding center for the topic  closer to the new label word, in this case, \textit{India}.}
    \label{fig:embeddings}
\end{figure*}

This section reviews topic models: how they are useful to
practitioners, the shortcomings of probabilistic and neural topic
models, and motivate our attempt to ameliorate this with
embedding-based interactions.

\subsection{Latent Dirichlet Allocation} \label{lda}

Topic models are exemplified by latent Dirichlet
allocation~\abr{lda} \citep{blei-03}. 
Given a large collection of documents and an integer parameter~$K$,
topic models like \abr{lda} find the $K$ topics that best describe the
collection.

\abr{lda} posits a generative story for how the data came to be and
uses probabilistic inference to find the best explanation for the
dataset~\citep{griffiths-04}.
While we do not fully recapitulate the \abr{lda} generative story
here---our focus is on neural models after all---the key is that one
part of the story is a distribution over words for each of the $K$
topics.
Often, one of the first steps of using the output of a topic model is
to \emph{name} the topics. Either by selecting top words through a
Markov chain Monte Carlo algorithm \citep{griffiths-topics,
hofmann-indexing} or through manual generation of descriptive
topics \citep{mei-spatio, wang-topics}. This is common especially in
the social sciences, where topics are given sentences to describe the
documents that make it up such as ``topic is associated with articles
on the life and works of Goethe''~\citep{riddell-12}.
For probabilistic models, however, this is not the end of the story.  
The Bayesian framework---through the use of informed
priors---encourages the incorporation of expert knowledge into interactive topic
models.
This can either represent a dictionary~\citep{hu-14}, word
lists from psychology~\citep{zhai-12}, or the needs of a business
organization~\cite{hu-14:itm}.
This feedback to a model helps correct word sense issues, match a
user's information needs, or reflect world knowledge and common sense.
Of course, one could move to a fully supervised
model~\citep{blei-07b}, where every training document has a topic label.
But this requires substantially more
interaction with the user than giving feedback on a handful of
topics---full supervision requires hundreds or thousands of labeled examples.
But these interactive models are not without their faults. 
First, they're slow; probabilistic inference---whether with \abr{mcmc}
methods or variational inference---struggles to update in the seconds
required to satisfy the best practices of an interactive application.
Second, while one of their goals is to incorporate the knowdge of
users, they completely ignore the vast world knowledge available ``for
free'' from representations trained on large text corpora.

\subsection{Neural Topic Models}

In addition to traditional topic modeling approaches like \abr{lda}, neural
topic models have emerged as a powerful alternative. These
models leverage deep learning techniques to capture complex relationships and
representations within textual data, offering several advantages over
traditional methods.
One of the key strengths of neural topic models is their proficiency in generating coherent and interpretable topics. This is primarily due to their use of nonlinear functions, which are more adept at closely matching the observed distribution of words and topics in the data. This nonlinearity allows neural models to capture complex and subtle semantic relationships, leading to topics that are not only semantically rich but also more aligned with human understanding. 

One popular architecture for neural topic modeling is the Variational Autoencoder (\abr{vae}) based topic model. \abr{vae}-based models, such as \abr{nvdm} encode documents into continuous latent spaces, enabling a smoother and more expressive representation of topics. For this model, we add topic embeddings directly to the latent space of the model, creating learnable topic embeddings similar to \abr{etm}. However, in contrast, \abr{nvdm} represents documents in a continuous latent space, unlike ETM and traditional models that use discrete distribution.

One of the primary advantages of neural topic models is their ability to generate more coherent and interpretable topics. Neural models can capture these nuances by learning distributed representations of words and topics. This leads to topics that are not only more semantically meaningful but also better aligned with human interpretations. \abr{etm} takes advantage of these representations by associating each topic with an embedding. These embeddings can be learned by the model or pre-trained word embeddings may be used. Like traditional topic models, each document has a vector connecting it to the $K$ latent topics. While a traditional topic model would have a full distribution over the vocabulary, in ETM the $k^{th}$ topic is a vector $\alpha_k$ $\in$ $R^L$---just like words in the embedding space. ETM induces a per-topic distribution over the vocabulary from that representation.

Moreover, they can handle large-scale text corpora efficiently and can adapt to different domains and languages or utilize the knowledge of large language models (\abr{llm}). In \abr{ctm} researchers sought to leverage the knowledge of \abr{llm} for better word representations. One such method is to combine the traditional \abr{bow} method with word embeddings from a \abr{llm} to develop contextualized embeddings which lead to better topic models \citep{bianchi-ctm}. The symbolic meaning of traditional \abr{bow} is lost after a single neural layer, so they hypothesize that contextualized embeddings would improve this, resulting in a stronger neural topic model. As \abr{ctm} are one of the best neural topic models, we decide to extend \abr{ctm} to be interactive as well and use it as our focus for our human study. 

%% file: sections/30-methods.tex
%

\begin{table*}[th]
    \centering
    \begin{tabular}{c c c c}
         & Vocab Size & Coherence & Diversity\\
         \hline
         \multirow{3}{4em}{ETM} & 2565 & 0.19 & 0.81 \\
         & 3572 & 0.17 & 0.85 \\
         & 10830 & 0.11 & 0.92\\
         \hline
         \multirow{3}{4em}{I-NTM (ETM)} & 2565 & 0.14 & 0.84\\
         & 3572 & 0.10 & 0.88\\
         & 10830 & 0.10 & 0.94 \\
         \hline
         \multirow{3}{4em}{I-NTM (CTM)} & 2565 & \textbf{0.21} & 0.91\\
         & 3572 & 0.18 & 0.92\\
         & 10830 & 0.15 & \textbf{0.95} \\
         \hline
    \end{tabular}
    \caption{Interactivity improves downstream classification tasks and the overall diversity of topics. In some cases, topic coherence decreases since coherence improves with general topics and we are labeling topics. Topic coherence and topic diversity, varying vocabulary sizes for \abr{etm} and various \abr{i-ntm} models on the \abr{better} dataset. Our both models under \abr{i-ntm} outperform standard \abr{etm} in terms of topic diversity and topic coherence }
    \label{tab:acc_tab}
\end{table*}

In this section, we explore the rationale and methodology behind modifying labels in neural topic models, focusing on two primary mechanisms: learnable topic embeddings and topic embeddings added in post-training. These methods offer similar, yet distinct approaches to refining topic models.
Traditional topic models often suffer from the absence of explicit labels, leading to potential mismatches between documents and topics or the generation of incoherent topics. This can lead to
situations where documents are associated with topics that they should
not be~\citep{ramage-etal-2009-labeled} or topics that just do not make
sense~\citep{newman-aetc}. Also, they require users to manually analyze
the topics found and then use labels such as the \underline{Business} topic.
Non-technical users also use a similar process when
using topic models: they inspect the topics, find the topics relevant
to their use case, and label them accordingly.
Thus, since labeling is a natural way people have already been interacting with topic models, we use labeling to both improve topics and help guide the
model to relevant topics for the users.

For instance, in a humanities context, instead of retrospectively describing topics with sentence-long descriptions, specific labels like `works of Goethe' can actively improve the model's relevance and accuracy.
Topic models can be used in time-sensitive situations such as
identifying key areas that need relief supplies through social media
postings ~\citep{resch-disaster, zhang-tm-disaster}.

Let's say a virus outbreak has occurred and a user wants to track what countries in Asia are currently affected. Given a large corpus this would be laborious to find the answer, so they turn to topic modeling. However, the user might already have an idea on what would be relevant to the questions being asked. With \abr{i-ntm} a user could label the topics as they deem fit and find relevant documents quicker, as in Table~\ref{tab:example}.  
\footnote{The scenario and
scenario specific questions, as well as the data used is from a dataset that focuses on
disaster relief situations \citep{mckinnon-rubino-2022-iarpa}.}

We will dissect two key methods in \abr{i-ntm} for updating topics, depending on the underlying model used. The first method involves learnable topic embeddings, that is models that have or can have learnable topic embeddings.
The second method is post-training adjustments, where topic embeddings are added in and modified after the model has been trained. 
These methods offer a suite of neural topic models to use interactively. By combining learnable topic embeddings and post-training adjustments, I-NTM provides a robust and flexible framework for users to interact with and guide the development of neural topic models.

\begin{figure*}[th]
    \centering
    \includegraphics[width=\textwidth,]{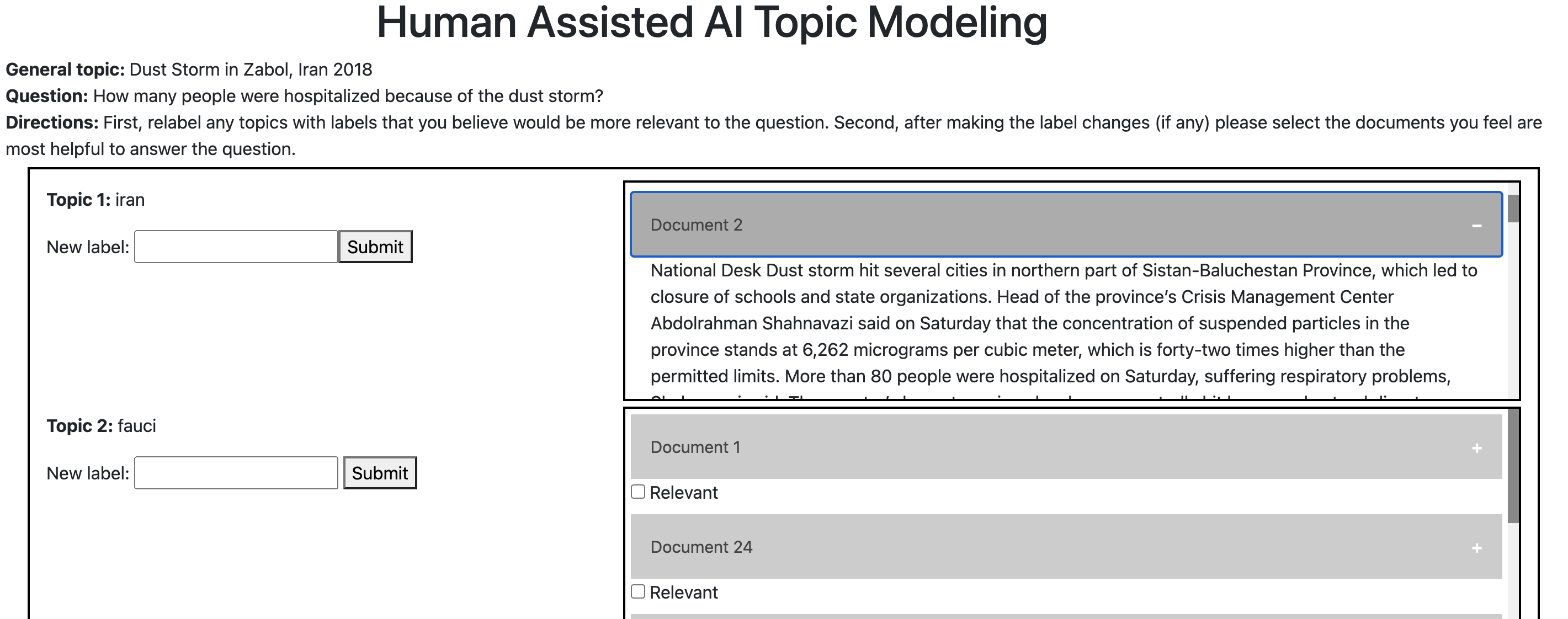}\hfill
    \caption{Human study interface for \abr{i-ntm}, using \abr{ctm} as the neural model. Users can see the
      given topics that are found for a set of tasks/requests and can
      change the label to better fit their needs. Additionally, the
      assigned documents for each topic are shown and users can select
      which documents are most relevant.}
    \label{fig:interface}
\end{figure*}

\subsection{Adjusting Learnable Topic Embeddings} \label{section:adjust}

In this section, we explore the first of two primary methods for updating topics in neural topic models: models that have or can have learnable topic embeddings.
As discussed above, for \abr{etm} and \abr{nvdm} we induce a topic
distribution from word representations and a topic embedding. These models
fall under the type of neural models where topic embeddings are or can be
directly represented in the model and therefore changed.
%
To make the topic modeling interactive, we allow for the users to
adjust the underlying embedding for each topic, thus ``moving'' the
topic closer to the word embeddings they desire.
We will discuss what this looks like in terms of users' actions in a
moment, but for the moment we assume that this can be expressed as a
vector
\begin{equation}
\Vec{\alpha_k}^{new} = \lambda(\Vec{w_k} - \Vec{\alpha_k}^{old}) + (1-\lambda) \Vec{\alpha_k}^{old}
\label{eq:intm-eq}
\end{equation}
where $\alpha_k^{old}$ is the topic embedding generated by the model
and $w_k$ is the word embedding associated with the topic the user
inputs.
That is, if the user wants a topic of \underline{food}, the topic
embedding is moved toward the word embedding corresponding to
\underline{food}.
The weight of adjusting the topic embedding towards the new label, can
be tuned through the parameter $\lambda$, which determines how close
the topic embedding is moved.

Following the example in
Table~\ref{tab:example}, Figure~\ref{fig:embeddings} shows the topic and word embeddings
before and after the adjustment of \underline{Topic~2}. The
words surrounding \underline{Topic~2} before adjusting the label, do not at first seem to be relevant to the question. After labeling of \underline{Topic~2}, as India, we see the topic embedding, is
close to the words ``india'', ``denv'', and ``scientist'', which are more likely to be relevant to the question and to reveal more relevant documents. 
\subsection{Adding Adjustable Topic Embeddings After Training}
\abr{etm} and \abr{nvdm} have trainable embeddings, but what about models that
cannot or adding them negatively affects training? In such cases, the idea is to introduce a form of topic embedding post-training, to enhance the model's performance and interpretability. We can simulate the effect of an embedding by creating a weighted average over the words that constitute a topic. This weighted average essentially serves as a stand-in for a physical topic embedding.
Our interactive framework supports these types of models. In this case,
the topic embedding is a proxy for a physical embedding to change the
topics. \abr{ctm} falls into this category.
Here, given a new label $w_l$ for a topic $t_i$, the distribution over words
for $t$ is updated to have higher probability for $w_l$ and for similar words,
$w_s$:
\begin{equation}
P_{\text{update}}(w_l \g t_i) = P_{\text{original}}(w_l \g t_i) + \Delta P(w_l \g t_i)
\end{equation}
and for similar words,
\begin{equation}
\Delta P(w_s \g t_i) = \lambda \cdot \text{similarity}(w_l, w_s) \cdot \Delta P(w_l \g t_i)
\end{equation}
where $\Delta P(w \g t)$ is the amount by which you increase the probability of word, $w$, in topic, $t$.

In neural topic models, topics are typically represented as distributions over words. Each topic is a blend of various words, with certain words having more weight or influence in defining the topic.
Thus, when a user assigns a label to a topic, they are providing a semantic point of reference for that topic. The model is prompted to adjust the weights of words in the topic distribution to align more closely with the semantics of the label. 

\subsection{User Interface}

While Equation~\ref{eq:intm-eq} outlines a theoretical framework for labeling topics, its practical application hinges on a user-friendly interface that allows for real-time interaction. To address this, we have developed an interface, as depicted in Figure~\ref{fig:interface}, which not only makes interactive topic modification feasible for neural models but also enhances user engagement beyond existing \abr{ntm} visualizations.
Our interface is designed with low-latency interactions in mind, a crucial feature for ensuring efficient topic refinement. There is immediate feedback when users label or re-label topics, fostering a dynamic interaction where users can intuitively understand the impact of their inputs on the model.

Furthermore, the interface is tailored to accommodate users without technical expertise. It allows them not only to assign labels to topics but also to observe, in real-time, how such labeling alters the document-topic assignments. This level of interaction is an advancement over traditional \abr{ntm} visualizations, which typically offer static or less responsive user experiences.
Users can delve into the topics, peruse associated documents, and input new labels via the interface. 

The underlying system seamlessly handles the complex tasks: adjusting topic embeddings, recalculating document-topic distributions, and updating the display to reflect these changes. This backend processing ensures that the interface remains user-friendly and effective.
A key feature of our interface is its ability to support continuous topic updates. Users can modify a topic multiple times, and there is flexibility to update several topics concurrently. To maintain the coherence and distinctiveness of topics, the interface incorporates safeguards against creating duplicate topics or topics with terms not found in the existing vocabulary.

\begin{figure*}
    \centering
    \includegraphics[width=\columnwidth]{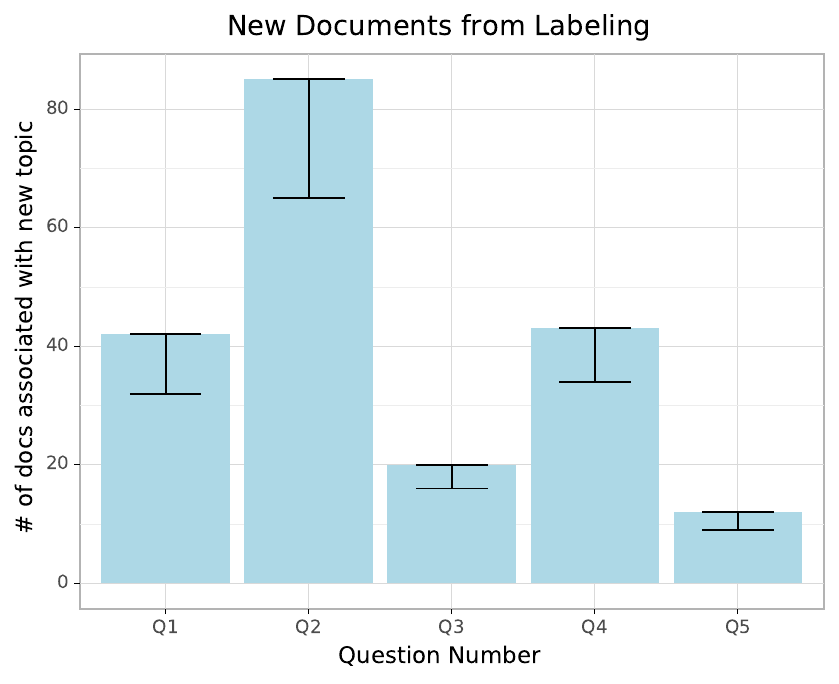}
    \caption{Labeling topics leads to, otherwise missed, documents to be revealed. The maximum number of new documents, that is, a document that was not previously associated with the topic, found for each question across all users. A single labeling of a topic can lead to a large number of new documents to be revealed.}
    \label{fig:study_plot}
\end{figure*}

\subsection{Automatic Metrics}
Since \citep{Dieng2020TopicMI} showed improvements in topic coherence
and diversity over \abr{lda}, to check if our method negatively
affects them, we looked at coherence and topic diversity for
varying vocabulary sizes between the different \abr{i-ntm} models.
Topic coherence is an automated method for evaluating the semantic
similarity of top words in a given topic. We measure the
normalized Pointwise Mutual Information (\abr{npmi}). Given a set
vectors of the top words in a topic, \{$w_1, w_2, ..., w_N$\}, the
\abr{pmi} is $$\text{PMI}(w_i, w_j) = \log_2
\frac{p(w_i, w_j)}{p(w_i)p(w_j)}.$$ \abr{npmi} is just an extension of
\abr{pmi}, where the vectors are weighted
\citep{aletras-stevenson-2013-evaluating}.

For our user study, we use information retrieval (\abr{ir}) as a metric for evaluating human labeling. \abr{ir} focuses on retrieving documents that are relevant to a given query. By using \abr{ir} as a metric, one can objectively assess how well user labeling improves the model's ability to retrieve relevant documents. If the user-labeled topics lead to more relevant documents being retrieved in response to a query, this indicates that the labeling process is effective. 
Additionally, \abr{ir} focus on retrieving relevant documents mirrors real-world use cases of topic models. By using \abr{ir} as an evaluation metric, you ensure that your assessment reflects practical scenarios where users rely on the model to find information quickly and accurately.

\subsection{Human Study}

To validate the efficacy of \abr{i-ntm}, we recruit participants to test our
model in finding more relevant documents for different information needs,
ex. documents that relate to foreign intervention in Cuba. Information
retrieval tasks are an intuitive way to measure the success of our method,
since they involve finding relevant information specific to a need. To verify
that user labeling uncover more relevant documents, we compare document
ranking scores before and after labeling and between a control group, where no
labeling is done and the test group, where users can label topics. Users that
label topics find more relevant documents that improve ranking scores more quickly.
\textbf{Setup}
We recruited 20 participants through the online platform Prolific.
\begin{enumerate*}
    \item Our model \abr{i-ntm} generates topics on the Text REtrieval Conference (\abr{trec}) Question Classification dataset. We randomly selected approximately 1500 documents from the Foreign Broadcast Information Service (\abr{fbis}).
    \item Participants see an information need with topics generated by our model. They can label topics as they deem best
    \item After labeling topics, they select a maximum of five documents that they believe best answer the information need 
\end{enumerate*}
We limit users to five minutes per question and for the users to select five
documents to normalize the results across all users and limit outliers,
i.e. a user taking an hour to comb through hundreds of documents to achieve
maximum ranking score.
Additionally, we want to mimic real-world scenarios
where thousands of documents and possibly hundreds of questions need to be
answered, so users would not have time to spend hours on each question.  For
each user we collect the topic information and document distribution before
and after the human interaction. Then using the algorithm commonly known as
BM25 \citep{Robertson1994OkapiAT}, an information retrieval ranking function,
we compare the estimated relevancy of topics before and after the human
interaction. We use BM25 since no gold relevance annotations are available for
the TREC dataset and since BM25 works by using a bag-of-words retrieval
function that ranks a set of documents based on query terms present in the
document, this is an effective way to compare retrieval performance between
our two groups.

%% file: sections/40-experiments.tex
\begin{table*}[th]
    \centering
    \begin{tabular}{c  c  c  c}
         \hline
         General Topic & Type & Avg Time Taken & Avg Number of Documents Selected \\
         \hline
         \multirow{2}{2em}{Cuba} & Control & $5^*$ min & 3 \\
         & Interactive & 2 min & 5 \\
         \hline
         \multirow{2}{2em}{South Korea} & Control & 5 min & 3 \\
         & Interactive & 4 min & 5 \\
         \hline
         \multirow{2}{2em}{Taiwan} & Control & $5^*$ min & 3 \\
         & Interactive & 2 min & 5 \\
         \hline
         \multirow{2}{2em}{Yugoslavia} & Control & $5$ min & 3 \\
         & Interactive & 3 min & 5 \\
         \hline
         \multirow{2}{2em}{China} & Control & 5 min & 5 \\
         & Interactive & 4 min & 5\\
         \hline
    \end{tabular}
    \caption{Our interactive method led to document selection, with more relevant documents being selected, on average. For the 5 different questions, the general topic of that question, the average amount of time a user spent on each question, and the average number of document selected are reported. A time of $5^*$ indicates they hit the set time limit of 5 minutes per question.}
    \label{tab:time-table}
\end{table*}
We evaluate \abr{i-ntm} on standard evaluation metrics and through a human study. Our experiments confirm that an interactive topic modeling interface greatly improves users' ability to find relevant documents in a timely manner. 

\subsection{Labeling Improves Coherence}\label{section:4.1}

Initially, we tested \abr{i-ntm} with automatic metrics without any human intervention, to understand how interaction changes the coherence and diversity of topics. Using \abr{etm} backend, topic coherence drops with our method, but diversity is
higher (Table~\ref{tab:acc_tab}). This effect is 
dataset dependent. For Wikipedia, adjusting six of
the topics to have distinct labels for
classification results in a more diverse topic
words. However, coherence typically improves with more general
clustering topics, since it measure co-occurence of words in the
documents with the topic words. So, with distinct topics, this can
result in lower topic coherence.
In contrast, the documents in the \abr{better} dataset (Table~\ref{tab:example} and Figure \ref{fig:embeddings}) are curated to be
related to disaster situations.
In this case, when topics are labeled to better fit the request at
hand, the topic words tend to have more overlap, since the request is
so specific. With the \abr{better} dataset, \abr{i-ntm}
decreases topic diversity but increases in
topic coherence.

Regardless, topic coherence is an imperfect metric for neural topic
modeling evaluation \citep{hoyle-21}. Nevertheless, we report these
scores for coherence and diversity since this is the current standard
for topic model evaluations. 

Human validation is viewed as the gold standard when it comes to topic model evaluation, thus we report those results in the next section.

\subsection{Human Study}\label{section:4.2}

\begin{figure}[t]
    \centering
    \includegraphics[width=\columnwidth]{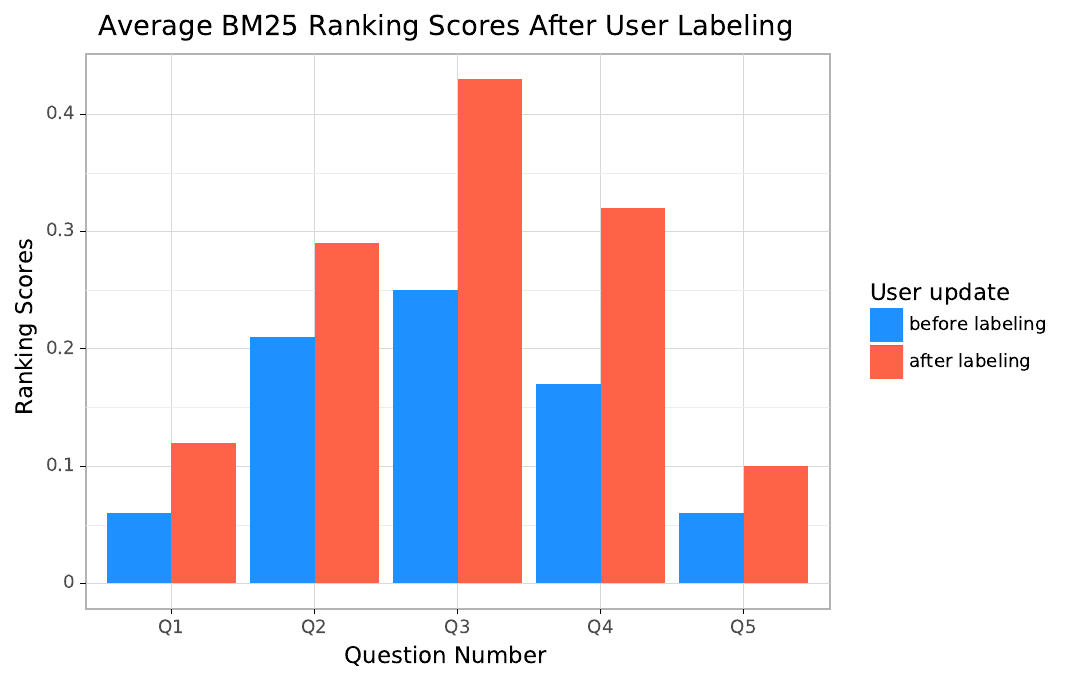}
    \caption{Average BM25 document ranking scores for each of the 5 questions averaged, over the 20 users. User inputted topic labels find more relevant documents and significantly improve document ranking scores}
    \label{fig:study}
\end{figure}

To evaluate the effectiveness of our interactive topic model interface, we conducted a human study using both a control (no labeling) and interactive (allows for labeling) scenario. For both treatments, the \abr{ctm} neural model is used. The control group and interactive group are given the same question, set of documents, and topic model. However, the control group is asked to find relevant documents without the ability to label any topics. In contrast, the interactive group can label topics and then select relevant documents.
When comparing the BM25 document ranking scores of the control and interactive group, we find labeling topics leads to more representative documents being revealed and chosen, when averaged across all 20 users (Figure~\ref{fig:study}).

In all cases except for Q5, there is a stark increase in ranking scores after the update. Question 5 was \textit{``Find document related to Chinese economic intervention in other countries''} and due to a large portion of the documents related to or came from Chinese news sources, it seemed easy for users to find documents discussing China's economic relations in comparison to the other questions. We see further evidence of this in Table~\ref{tab:time-table} where Q5 had the highest average number of documents selected across all control scenarios and matched the interactive scenario in average time taken. In contrast, the other four questions took more time and selected less documents than the interactive case. In the case of Q1 and Q3 where the time limit was reached, the users were not able to find 5 related documents in time.
While a high number of new associated documents does not necessarily correlate with an increase in document ranking score, as some of the new documents might be related to the general topic but not the specific information need, we find that labeling does reveals a significant amount of documents that were not previously in that topic (Figure \ref{fig:study_plot}). Again, we see a significantly smaller number for Q5, which we believe is due to the prevalence of documents related to China in the dataset.

The \abr{trec} dataset contains a large amount of global affairs documents, with many mentioning multiple topics. So while the \underline{Cuba} topic found by the topic model will contain documents relevant to Cuba, for the question of \textit{"Find documents related to Cuba and US relations"}, the users have to filter through many unrelated documents. In this case, a label brought forth more relevant documents. 

%% file: sections/60-related.tex
Topic modeling covers a wide range of methods for discovering topics within a
corpus and there has been extensive research across these different
methods. We discuss these similar methods and contrast them with our own in
the following seciton.

\paragraph{Neural topic models}
With the recent developments in deep neural networks (\abr{dnns}, there has been work to use these advancements to increase performance of topic models. One of the most common frameworks for neural topic models (\abr{ntms}), described in \citep{zhao2021topic}, as \abr{vae-ntms}. Much research was focused on adapting \abr{vae}'s for topic modeling; \citep{zhang-whai, srivastva-vaenmt} focus on developing different prior distributions for the reparameterization step of \abr{vae}, such as usung hybrid of stochastic-gradient \abr{mcmc} and approximating Dirchelt samples with Laplace approximations. \abr{vae}-\abr{ntm} also were extended to work with different architectures, \citep{nallapti-sengen} developed a sequential \abr{ntm} where the model generates documents by sampling a topic for one whole sentence at a time and uses a \abr{rnn} decoder. \abr{etm} and therefore, \abr{i-ntm} use these advancements in \abr{vae} to update the neural model parameters.

\paragraph{Interactive topic modeling.} Interactive labeling of topics has been thoroughly explored for probabilistic topic models. Works involving labeling topics through images using neural networks, using a sequence-to-sequence model to automatically generate topics, or using unsupervised graphical methods to label topics~\citep{aletras-labeling, aletras-stevenson-2014-labelling, aletras-auto}.
\citep{pleple-13} designed an interactive framework that allows the user to give live feedback on the topics, allowing the algorithm to use that feedback to guide the \abr{lda} parameter search.
\citep{smith-etal-2017-evaluating} compared labels generated by users after seeing topic visualizations with automatically generated labels. 
\citep{hu-14:itm} provides a method for iteratively updating topics by enforcing constraints.
\citep{mei-auto-topic} make the task of labeling into an optimization problem, to provide an objective probabilistic method for labeling. But there has yet to be work that extends this iterative process to neural-based topic models in an intuitive and natural sense such as \abr{i-ntm}. 
There has been extensive work in the area of anchor-based topic modeling---where a single word is used to identify a topic. \citep{Lund-17} present ``Tandem Anchors'' where multi-word anchors are used to interactively guide topics. \citep{Yuan2018MultilingualAI} developed a framework for interactively establishing anchors and alignment across languages. \citep{dasgupta-int} introduces a protocol that allows users to interact with anchor words to build interpretable topic. 
The most similar and recent work to outs is \citealp{fang-etal-2023-user} which simultaneously developed a user-interface for interactive and guided topic modeling, based on Gibbs sampling. While it has obvious similarities, we developed the first interactive interface for neural topic models and have an interface that users can see in real-time their changes to the model.   

\paragraph{Automatic topic modeling} For a similar purpose, but through a different process, many works have sought to automatically generate labels. \citep{aletras-auto} where they re-rank labels from a large pool of words to label topics in a two-stage method. \citep{lau-11} uses top terms from titles and subwords from Wikipedia articles to rank and label topics based on lexical features. \citep{Mao-12} exploit the parent-sibling relationship of hierarchical topic models to label the topics. Unsupervised methods that differ from topic models but with the same goal of clustering data also exist. \abr{llm} can be prompted to cluster data with or without labels in an intelligent way \citealp{wang-etal-2023-goal}

\nocite{Yuan2018MultilingualAI,Yang2019AMT,Yuan2020InteractiveRO,Dieng2020TopicMI}

%% file: sections/70-conclusion.tex
%

We introduce \abr{i-ntm}: a method and interface for users to interactively update topics given by neural topic models. While there have been previous efforts to improve probabilistic topic modeling through labeling, this is the first work to our knowledge that allows interactive updating of neural topic models to improve the found topics. Especially in real-world situations, such as disaster relief, the ability to improve topics through labeling allows non-technical users to tailor the topics to their specific needs.

Additionally, our user study verifies that giving users the ability to label topics improves performance on downstream information retrieval tasks in less time, validating that more relevant documents are being found. 

To take this work further and give as much flexibility to the user as possible adding the ability to guide the training of topic models by interactive labeling throughout the training, multi-word labeling instead of single, vocabulary based labels, stronger encoders, and direct access to making the adjustments in the embedding space through embedding visualizations would improve upon this presented method. Finally, while we present a suite of three different neural models, interactive topic modeling and by extension our interface, could be extended to other models such as LLMs..

%% file: sections/80-limitations.tex
This work we seeks to solve a key limitation in traditional topic models--- guiding the topics of a model in a way that is relevant to the user. 
Along the lines of what it means to ``help'' identify more relevant topics, \citep{hoyle-21} discusses the limitations of coherence, an automatic metric for topic model evaluation. Topic coherence is an automatic metric that is not validated by human experiments and thus its validity of evaluating topic models is limited. While our method is an attempt to improve interpretability of topic models, it still suffers from many of the problems that topic models in general do. Topic models do not conform to well-defined linguistic rules and due to the non-compositionality of labels, from a linguistic viewpoint, can be viewed as not actually modeling topics \citep{shadrova-1}. 

We recognize that with any study there are limitations, while topics are meant to be representative labels of the corpus, users tended to use words directly in the query or general task, treating it more as a keyword match. While this is not how topic models are meant to be used and most likely due to a lack of knowledge about topic models, this process did work in most cases at improving the relevancy scores for the questions. 

Finally, the BM25 requires a query to calculate the scores. We used the scenario and corresponding question as the query (removing stopwords), however a variation in query could lead to different BM25 scores. While this does not change the fact that labeling topics on average improved BM25 scores, it means a good query is required to effectively rank documents.

%% file: sections/90-ethics.tex
The data that we used for the experiments in this paper was all human gathered by others and ourselves. If \abr{i-etm} was to be used in a real-word situation, where identifying key documents or tweets about a time-sensitive issue was paramount, any failures in the system could result in a negative outcome if the wrong information is disseminated. We went through the appropriate IRB pipeline to receive approval for our human conducted study. The users were paid based on the recommendation of the Prolific platform, which bases its' recommendation based on the time of the study and other studies. This was a rate of $\$12$ an hour. No personal identification information was collected from the users, so there poses no threat to the participants of exposure of personal information.

%% file: sections/100-appendix.tex
We used the BETTER dataset and a curated Wikipedia dataset.\footnote{https://github.com/forest-snow/mtanchor\_demo} To preprocess the data, we removed English stopwords and used the 0.01 and 0.85 as the minimum and maximum document frequency, respectively.

\subsection{Training details}
For all the results presented in this paper, our model was trained using 4 NVIDIA RTX2080ti The \abr{i-ntm} model was trained for 200 epochs using 20 topics. The ADAM optimizer is used with a learning rate of 0.005. \footnote{we followed the other default parameters in the original paper and can be found in our code as well.} The rest of the details can be found in the appendix. For our human study, we trained a model using only 5 topics. This was due to not wanting to overwhelm users with a lot of topics and the limited number of documents in the dataset.

\section{Models}
We used the PyTorch implementation of ETM to build our code off of \footnote{https://github.com/lffloyd/embedded-topic-model}. We used an embedding space size and rho size of 300 and a hidden layer size of 800. The rest of the hyperparameters are the default and can be found in the original code or our own. To greatly improve training time, we used the pre-trained fasttext embeddings \cite{mikolov2018advances}.

\section{Code}
The code will be publicly made available on our Github page.